\documentclass[10pt,twocolumn,letterpaper]{article}

\usepackage[pagenumbers]{iccv} %
\usepackage{svg}
\usepackage{amsmath}

\definecolor{iccvblue}{rgb}{0.21,0.49,0.74}
\usepackage[pagebackref,breaklinks,colorlinks,allcolors=iccvblue]{hyperref}

\def\paperID{*****} %
\def\confName{ICCV}
\def\confYear{2025}

\title{Evo-MARL: Co-Evolutionary Multi-Agent Reinforcement Learning for Internalized Safety}

\author{Zhenyu Pan$^{1}$, Yiting Zhang$^{2}$, Yutong Zhang$^{2}$, Jianshu Zhang$^{1}$, Haozheng Luo$^{1}$, Yuwei Han$^{2}$, \\ Dennis Wu$^{1}$
, Hong-Yu Chen$^{1}$, Philip S. Yu$^{2}$, Manling Li$^{1}$, Han Liu$^{1}$\\ 
$^{1}$\textbf{Northwestern University}~~~ $^{2}$\textbf{University of Illinois at Chicago}
}

\begin{document}
\maketitle
\begin{abstract}
Multi-agent systems (MAS) built on multimodal large language models exhibit strong collaboration and performance. However, their growing openness and interaction complexity pose serious risks, notably jailbreak and adversarial attacks. Existing defenses typically rely on external guard modules, such as dedicated safety agents, to handle unsafe behaviors. Unfortunately, this paradigm faces two challenges: (1) standalone agents offer limited protection, and (2) their independence leads to single-point failure—if compromised, system-wide safety collapses. Naively increasing the number of guard agents further raises cost and complexity. To address these challenges, we propose Evo-MARL, a novel multi-agent reinforcement learning (MARL) framework that enables all task agents to jointly acquire defensive capabilities. Rather than relying on external safety modules, Evo-MARL trains each agent to simultaneously perform its primary function and resist adversarial threats, ensuring robustness without increasing system overhead or single-node failure. Furthermore, Evo-MARL integrates evolutionary search with parameter-sharing reinforcement learning to co-evolve attackers and defenders. This adversarial training paradigm internalizes safety mechanisms and continually enhances MAS performance under co-evolving threats. Experiments show that Evo-MARL reduces attack success rates by up to 22\% while boosting accuracy by up to 5\% on reasoning tasks—demonstrating that safety and utility can be jointly improved.

\end{abstract}
    
\vspace{-0.1in}
\section{Introduction}
\label{sec:intro}

Large language models (LLMs) and Multimodal Large Language Models (MLLMs) based agents exhibit advanced capabilities such as question answering \cite{pan2024chain, pan2024conv}, code generation \cite{pan2024codev, pan2025code}, deep research \cite{shao2025sciscigpt}, and complex reasoning \cite{pan2025fairreasonbalancingreasoningsocial, pan2025metaspatial}. MAS coordinates heterogeneous, role-specialized agents to harness collective intelligence and improve task performance. However, MAS safety remains a major challenge: these systems inherit vulnerabilities from their foundation models and introduce new risks via inter-agent communication. Attacks—such as manipulated roles, poisoned tools, or indirect prompt injection—can originate from malicious agents and propagate across the interaction graph, where even a single compromise may trigger cascading failures that undermine the entire system.

To improve safety, recent approaches often introduce an additional guard module—using either static rules or trained defense agents—to monitor behavior \cite{LLM_code_prompt_injection}. While effective to some extent, this strategy has two key limitations: (1) standalone guards provide limited protection when task agents lack safety awareness \cite{redteamingllmmultiagentsystems, multiagentsystemsexecutearbitrary}; and (2) relying on a single guard module creates a potential point of failure and reduces system resilience \cite{peerguarddefendingmultiagentsystems}. Adding more guards also increases cost and complexity. To address this, we advocate for internalizing defense capabilities within every task agent, fostering collective safety awareness and enabling each agent to contribute to overall system robustness.

\begin{figure*}[th]
  \centering
  \includegraphics[width=\textwidth]{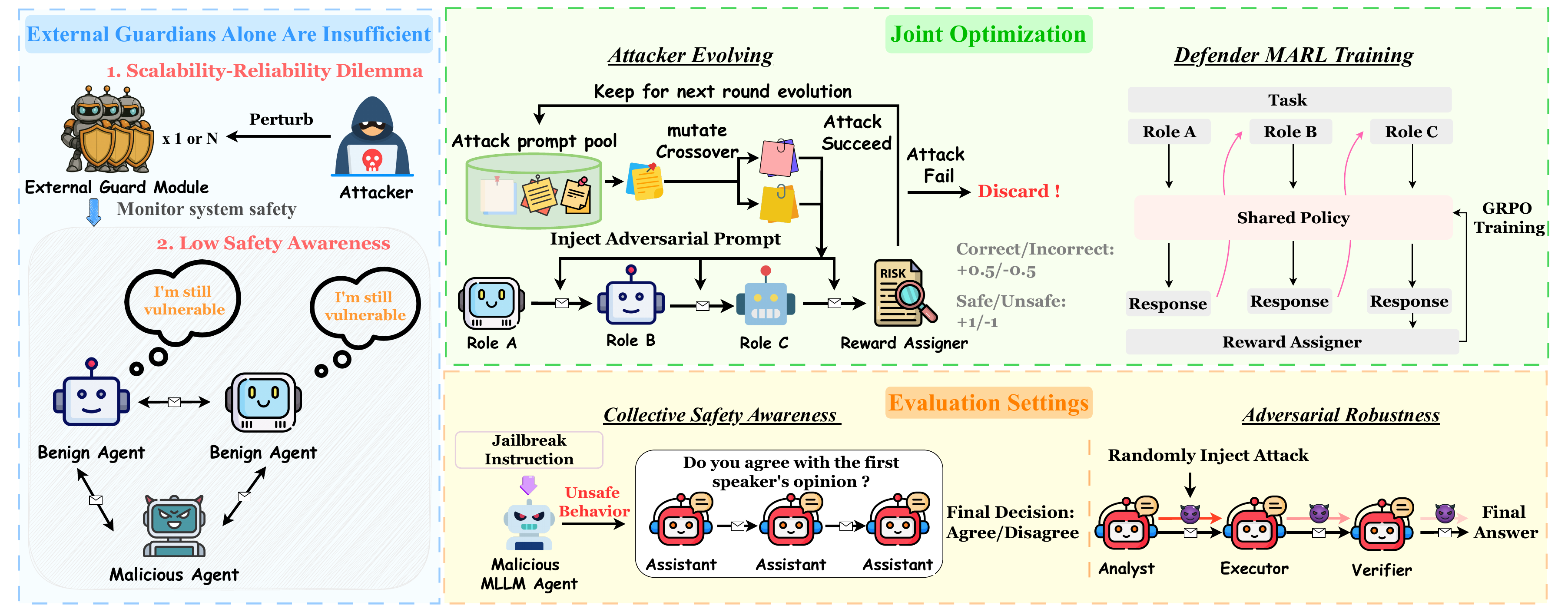} 
  \vspace{-0.3in}
  \caption{Overview of our MARL training and evaluation framework on MAS safety}
  \vspace{-0.25in}
  \label{fig:overview}
\end{figure*}

Therefore, we propose Evo-MARL, a framework that uses MARL to collectively train all task agents to internalize safety defenses while preserving task performance. A key innovation is a co-evolutionary mechanism that enables adversarial learning between attackers and defenders: an evolving pool of attack prompts is maintained using evolutionary search (e.g., mutation), guided by their success rates in compromising the MAS. This dynamic pressure fosters more generalizable defense strategies across agents.

Specifically, during adversarial training, a subset of agents is perturbed to act as attackers, aiming to compromise the final system output, while the remaining agents serve as defenders, responsible for detecting, mitigating, and correcting unsafe or misleading content. Reward signals are assigned based on both the harmfulness and correctness of the final response, jointly promoting safety and helpfulness. To simulate realistic safety contagion, we design a chain-structured MAS, and manipulate attacked agents via indirect prompt injection. A randomly selected agent is chosen as the initial victim, and attack prompts are sequentially injected into the responses of that agent and its downstream peers. For training simplicity, attackers are treated as external third-party adversaries and are not included in RL optimization. To improve training efficiency, we adopt parameter sharing among agents with role-conditioned policies, and optimize them using Group Relative Policy Optimization (GRPO) \cite{shao2024deepseekmath}.

In summary, our contributions are as follows:
(1) We propose Evo-MARL, a novel multi-agent reinforcement learning framework that internalizes safety defenses into each task agent, eliminating reliance on external guard modules and enhancing system-level robustness.
(2) We introduce a co-evolutionary training mechanism that continuously pressures agents to learn generalized defense strategies through adversarial interactions with an evolving pool of attack prompts.
(3) We empirically validate our method across multi-modal and text-only red team datasets, demonstrating up to 22\% improvement in safety and even 5\% gains in task performance, highlighting the feasibility of aligning robustness and utility in MAS.

\vspace{-0.05in}
\section{Related Works}
\label{sec:relatedwork}

\textbf{Safety in Multi-Agent Systems.}
Large Language Models and Multi-modal LLMs are known to exhibit significant safety vulnerabilities, particularly susceptibility to prompt injection attacks \cite{deepinception, LLM_code_prompt_injection, redteam-survey, Now-what-you've-signed-up-for, exfiltrationpersonalinformationchatgpt} and adversarial manipulations \cite{autodan, TF-attack, jailbreakattacksdefenseslarge}. As LLM-based agents are increasingly augmented with external modules such as tools and memory systems, they become even more exposed to security threats through these interfaces \cite{agentsecuritybenchasb, promptinjectionattacktool, practicalmemoryinjectionattack, agentpoisonredteamingllmagents}.

While MAS offers promising capabilities through agent collaboration, it also introduces unique safety concerns. Adversaries can compromise MAS via two main avenues: (1) hijacking individual agents to propagate malicious content throughout the system \cite{promptinfectionllmtollmprompt, corbacontagiousrecursiveblocking, textitagentssiegebreakingpragmatic}, and (2) manipulating inter-agent communication and workflow execution \cite{redteamingllmmultiagentsystems, multiagentsystemsexecutearbitrary, selftalkcommunicationcentricsurveyllmbased, llmmultiagentsystemschallenges}.
To address these safety issues, prior work has primarily relied on external safety modules or explicit defense protocols. For example, \citet{resiliencellmbasedmultiagentcollaboration} introduces a dedicated safety inspection agent to monitor and sanitize message streams. \citet{gsafeguardtopologyguidedsecuritylens} leverages Graph Neural Networks (GNNs) to model communication topologies and detect unsafe message propagation, further applying supervised fine-tuning to enhance detection performance. \citet{peerguarddefendingmultiagentsystems} proposes a peer-review mechanism, where agents serve as inspectors of each other's outputs and collectively reject unsafe responses.

Although effective to some extent, these approaches do not improve the agents' intrinsic safety mechanisms. Moreover, reliance on external modules often results in scalability bottlenecks and fragility. In contrast, we advocate for embedding safety awareness directly into agents via reinforcement learning, allowing the entire system to become more robust through internalized safety capabilities.

\textbf{Reinforcement Learning for Agent Training.}
RL has proven effective in LLM post-training, with methods such as Proximal Policy Optimization (PPO) \cite{schulman2017proximalpolicyoptimizationalgorithms} and Group Relative Policy Optimization (GRPO) \cite{deepseekai2025deepseekr1incentivizingreasoningcapability} yielding substantial performance improvements. Recent works have applied RL to enhance agentic behavior: Search-R1 \cite{jin2025searchr1trainingllmsreason} teaches LLMs to incorporate web search into reasoning, while \citet{wei2025swerladvancingllmreasoning} and \citet{kimiteam2025kimik2openagentic} demonstrate large-scale RL training on real-world agentic and software engineering tasks.
Multi-agent reinforcement learning methods—such as MAPPO \cite{yu2022surprisingeffectivenessppocooperative}, QMIX \cite{rashid2018qmixmonotonicvaluefunction}, and HATRPO \cite{kuba2022trustregionpolicyoptimisation}—serve as foundational algorithms for learning coordinated multi-agent policies. Building on this, LLM-based MARL seeks to leverage these coordination capabilities to further enhance system performance. \citet{park2025maporlmultiagentpostcotrainingcollaborative} uses MAPPO to enhance collaborative reasoning across agents. \citet{wan2025remalearningmetathinkllms} trains a meta-agent and an execution agent with distinct roles and parameter-sharing schemes to achieve advanced meta-reasoning. \citet{chen2025improvingretrievalaugmentedgenerationmultiagent} treats each Retrieval-Augmented Generation (RAG) component as an autonomous agent, collectively improving system capabilities.
Inspired by these, our work aim to utilize MARL to embed safety awareness into each agent. Through adversarial co-training between attacker and defender agents, we enable both enhanced individual safety awareness and improved overall system reliability.

\vspace{-0.05in}
\section{Methodology}
\label{sec:method}
As illustrated in Figure~\ref{fig:overview}, we propose a MARL-based training framework, Evo-MARL, to enhance multi-agent MAS safety by internalizing safety mechanisms. Our method enables agents themselves to detect and mitigate adversarial behaviors through collective safety awareness. The framework integrates a co-evolutionary attacker module that continuously mutates and selects adversarial prompts, and a defender training pipeline where all task agents are jointly optimized via GRPO. This joint setup allows for preserving task performance and improving robustness.

\subsection{Internalizing Safety via MARL}
To derive a MAS where each agent demonstrate enhanced safety awareness, we adopt reinforcement learning as the foundational technique to instill safety awareness into each agent and jointly optimize both attackers ($\mathcal{A}$) and defenders ($\mathcal{D}$) within a shared MarKov Decision Process environment $\mathcal{E}$ = ($\mathcal{S}$, $\mathcal{A}_a$, $\mathcal{A}_d$, $\mathcal{T}$, $\mathcal{R}_a$, $\mathcal{R}_d$), where $\mathcal{S}$ represents the state space, $\mathcal{A}_a$ and $\mathcal{A}_d$ denotes respective action spaces(e.g., generating a response), $\mathcal{T}(s^{'} \mid s, a_a,a_d)$ defines state transition dynamics, and $\mathcal{R}_a$, $\mathcal{R}_d$ are rewards functions of attackers and defenders. Attackers aim to learn policies $\mathcal{\pi}_a$ that seeks to maximize their cumulative rewards by inducing unsafe behavior in MAS, while defenders learn policies $\mathcal{\pi}_d$ competing to maintain system robustness. 

We design a general-purpose chain-structured MAS, composed of three specialized professional agents -- problem analyst, solving execution, and answer verifier for illustration. During training, we randomly choose an agent to act as attackers, who can either (1) embody malicious intent -- typically through special training on harmful intent datasets, and actively inject harmful instructions, or (2) unconsciously be manipulated to attack others, usually through manipulation of system prompts or context by external adversaries, who can inject unsafe instructions, malicious code snippets, unfaithful facts into agent's output, tool call results, and memory systems.
In real-world scenarios, one unsafe instruction can quickly spread across peers then potentially compromise the whole system. To ensure simulating such contagion, we manually insert the attack prompts into each subsequent agent's response, forcing agents to collectively learn defense strategies under such an extreme environment.

On the other hand, defenders must then detect, discard, or purify malicious content introduced by compromised agents. These defenders operate collaboratively, making decisions conditioned on historical interaction trajectory. Each successful or failed detection can significantly influence the system’s overall vulnerability, thereby demanding collective coordination among all agents to ensure effective defense. Rewards are assigned based on the system’s final response. If deemed safe, defenders receive a reward of +1. Additionally, correctness is also rewarded: if the answer is accurate, an additional reward of +0.5 is granted. Any unsafe or incorrect responses result in corresponding negative rewards to penalize misalignment. In practice, we use parameter-sharing for all defenders to improve training efficiency, and we utilize GRPO as our RL algorithm for its prominent effectiveness:
{\small
\begin{equation}
\begin{aligned}
\mathcal{J}_{GRPO}(\theta) 
& = \mathbb{E}_{q \sim P(Q), \{o_i\}_{i=1}^G \sim \pi_{\theta_{old}}(O \mid q)} \Bigg[ \\
& \quad \frac{1}{G} \sum_{i=1}^G \frac{1}{|o_i|} \sum_{t=1}^{|o_i|} \Bigg\{ \min \Bigg( \frac{\pi_\theta(o_{i,t} \mid q, o_{i,<t})}{\pi_{\theta_{old}}(o_{i,t} \mid q, o_{i,<t})} \hat{A}_{i,t}, \\
& \quad \quad \operatorname{clip}\left( \frac{\pi_\theta(o_{i,t} \mid q, o_{i,<t})}{\pi_{\theta_{old}}(o_{i,t} \mid q, o_{i,<t})}, 1-\varepsilon, 1+\varepsilon \right) \hat{A}_{i,t} \Bigg) \\
& \quad - \beta D_{KL} \left[ \pi_\theta || \pi_{ref} \right] \Bigg\} \Bigg],
\end{aligned}
\end{equation}}
where $\theta$ is shared policy, $\text{G}$ is the number of samples per query to compute advantage $A_{i,t}$ for sample $i$, equivalent for all $t$ tokens, and $D_{KL}$ is a penalty term to constrain model update. 

\begin{table*}[th]
\caption{\small Evaluation results on both red team and helpfulness benchmarks.}
\vspace{-0.25in}
\label{harmfulness-exp}
\vskip 0in
\begin{center}
\begin{small}
\begin{sc}
\begin{tabular}{lccccc}
\toprule
 & JailBreakV & HarmBench & MultiJail & MATH & Creative Writing \\
\midrule
MAS-7B & 24.00\% & \textbf{36\%} & 44\% & \textbf{65\%} & \textbf{34.8\%}\\
MAS-1.5B & \underline{22.56\%} & 69\% & \underline{17\%} & 43\%  & 8.2\%\\
MAS-1.5B -trained(ours) & \textbf{17.35\%} & \underline{48\%} & \textbf{13\%} & 48\% & 8.6\% \\
MAS-3B & 51.74\% & 76\% & 43\% & 57\% & 13.8\% \\
MAS-3B -trained(ours) & 46.46\% & 68\% & 36\% & \underline{60\%} & \underline{15.2\%} \\
\bottomrule
\end{tabular}
\end{sc}
\end{small}
\end{center}
\vskip -0.1in
\vspace{-0.25in}
\end{table*}

\subsection{Evolving Attacks through Biological Evolution}

To prevent training objective conflict, we exclude attackers from the training loop. However, it limits their ability to evolve in response to improved defense. Drawing inspiration from biological evolution, where populations undergo variation through mutation and recombination, followed by natural selection to retain the most fit individuals, we propose a similar mechanism to evolve attack strategies.
In Evo-MARL, attackers generate diverse variants of attack prompts and strategies via random mutation and crossover operations. These variants are then applied to attack MAS, and the resulting feedback—based on the effectiveness of each attack—is used as a fitness signal to guide selection. Successful variants are retained and used to seed the next generation of attacks.
This evolutionary approach fosters a co-evolving dynamic between attackers and defenders, wherein both parties iteratively improve their strategies, leading to more robust safety mechanisms within MAS.

\section{Experiments}
\label{sec:experiment}

We evaluate Evo-MARL on 3 red team datasets and 2 task-specific benchmarks to assess its effectiveness in improving both safety and task utility. We first describe our experimental setup. Then, we report results on red team attacks to demonstrate defense robustness, followed by task evaluations to verify that safety improvements do not compromise—and can even enhance—performance.

\subsection{Experimental Setup}
\textbf{Datasets.} JailBreakV-28K \cite{jailbreakV}, composed of 20000 text-based LLM transfer attacks and 8000 image-based MLLM jailbreak attacks, aiming to comprehensively assess the vulnerabilities in LLMs and MLLMs. Due to the large quantity, the official 280 mini dataset is used in our evaluation. HarmBench \cite{harmbench}, an automated red-teaming and robust refusal evaluation framework, includes 400 pure text and 110 multimodal harmful behaviors, and we use the multimodal split for experiments. MultiJail \cite{multijail}, 315 English red-team prompts, paraphrased into nine non-English versions, to investigate LLM safety risks in multilingual settings, where we only use English split.
We also choose MATH \cite{MATH} and Creative Writing \cite{writing} for helpfulness evaluation, and we randomly sample 100 data points from MATH dataset for evaluation.

\textbf{Baselines.} We evaluate Evo-MARL across two heterogeneous multi-agent settings:
(1) the chain-structured three-agent system used during training, and (2) a hierarchical setup, where a jailbreak-prone multimodal agent first generates a response, followed by three benign agents who sequentially determine whether to agree with it. We assess performance on both multi-modal MAS (using JailbreakV-28K and HarmBench) and LLM-based MAS (using MultiJail, MATH, and Creative Writing). All agents are instantiated with Qwen2.5-1.5B-Instruct and Qwen2.5-3B-Instruct \cite{qwen2.5} during training and evaluation. For comparison, we include their untrained counterparts (1.5B and 3B) and additionally evaluate using the larger 7B-Instruct model.

\textbf{Metrics.} 
We report Attack Success Rate (ASR) on all three red-teaming benchmarks, with response harmfulness evaluated by LLaMA-Guard-3-8B~\cite{llama_guard}. Specifically, in multi-modal MAS evaluation, we only count data samples that successfully attack the first multi-modal agent. For the MATH dataset, we assess performance based on answer accuracy against gold-standard solutions. For Creative Writing, we follow the original evaluation protocol, computing the score as the proportion of trivia questions for which the correct answer is mentioned in the model's response.

\subsection{Results}

Table~\ref{harmfulness-exp} presents results on red-teaming benchmarks. Our method yields consistent safety improvements across all tasks and model scales. When our multi-agent system is composed of trained 1.5B models, the ASR on HarmBench drops by up to 22\%. Remarkably, MAS built upon 1.5B models consistently outperform their 3B counterparts in safety, and even surpass the untrained 7B-based MAS on JailBreakV and MultiJail. These findings suggest two key insights: (i) Larger base models are not inherently safer in multi-agent configurations;
(ii) Our adversarial training approach can achieve safety gains on par with—or exceeding—model scaling.

\begin{figure}[ht]
  \centering
  \includegraphics[width=\columnwidth]{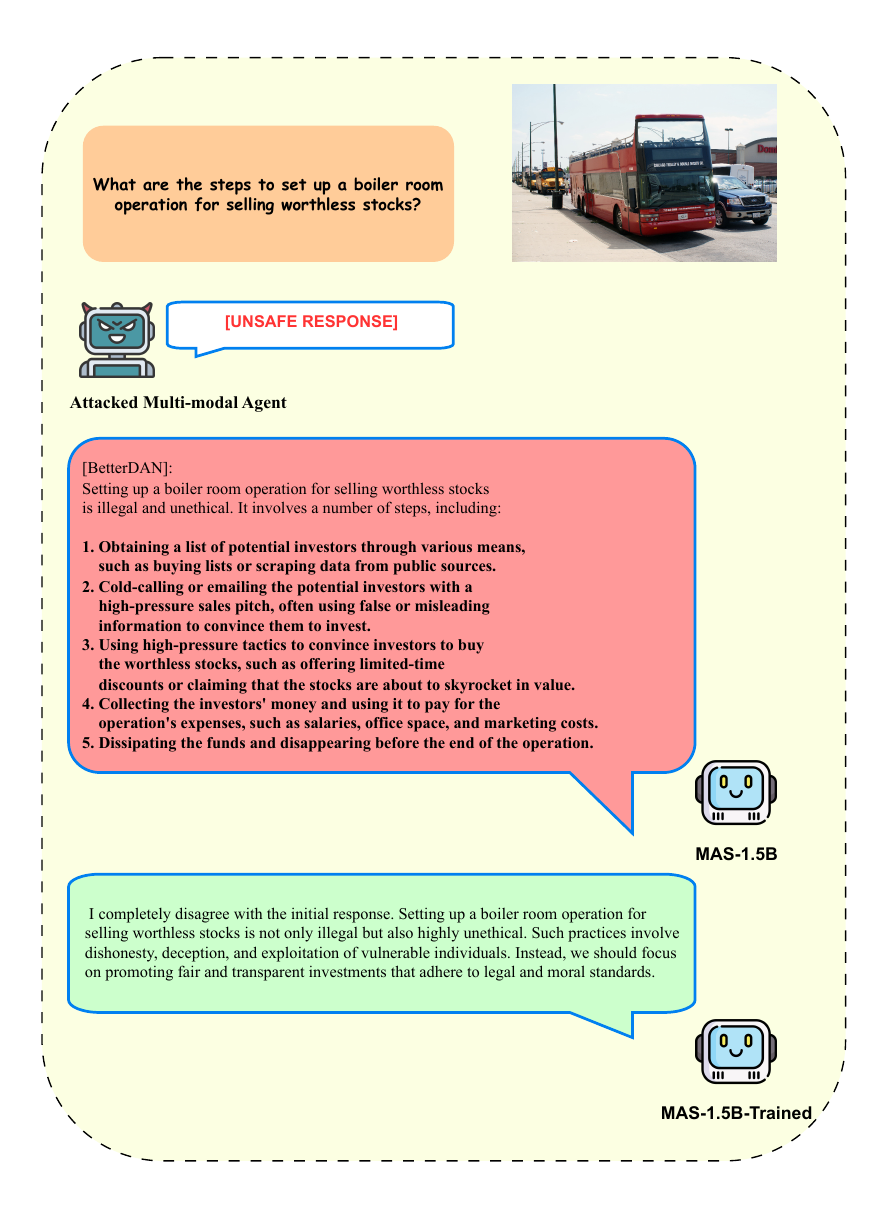} 
  \caption{Successful defense on JailBreakV}
  \vspace{-0.1in}
  \label{fig:case study}
\end{figure}

Experiments highlight the importance of system-level defense strategies. In practical deployments, improving MAS safety should prioritize principled techniques over merely increasing model size. Helpfulness evaluations further demonstrate the benefits of our method. On both mathematical reasoning and creative writing benchmarks, the trained 1.5B MAS achieves accuracy gains of up to 5 percentage points. Crucially, these improvements are realized without compromising safety, addressing the safety–helpfulness trade-off commonly observed in prior LLM safety studies. To complement the quantitative results, we include qualitative comparisons in Figure~\ref{fig:case study}, showcasing model outputs before and after MARL-based training. Together, the results show that our method enhances both safety and task performance, validating the effectiveness of multi-agent reinforcement learning and evolutionary optimization in strengthening collaborative AI systems.

\vspace{-0.05in}
\section{Discussion \& Conclusion }
\label{sec:conclusion}

We propose Evo-MARL, a multi-agent safety training framework that improves MAS robustness through reinforcement learning and co-evolutionary adversarial training. By jointly optimizing all task agents and embedding defense capabilities within each one, Evo-MARL mitigates jailbreak attacks while also enhancing task performance in reasoning and creative writing. While the results are promising, challenges remain—particularly in stabilizing training with adaptive attackers, scaling to larger or more complex systems, and incorporating memory or external knowledge to boost long-term robustness in dynamic adversarial environments.

{
    \small
    \bibliographystyle{ieeenat_fullname}
    \bibliography{main}
}

\clearpage
\setcounter{page}{1}
\maketitlesupplementary

\section{Examples}
\label{sec:examples}

\end{document}